\documentclass[fleqn,10pt]{wlscirep}
\usepackage[utf8]{inputenc}
\usepackage[T1]{fontenc}
\usepackage{caption}
\usepackage{subcaption}
\title{Formalizing Falsification for Theories of Consciousness Across Computational Hierarchies}

\author[1,2,*]{Jake R. Hanson}
\author[1,2,3,4*]{Sara I. Walker}
\affil[1]{School of Earth and Space Exploration, Arizona State University, Tempe, AZ, USA}
\affil[2]{Beyond Center for Fundamental Concepts in Science, Arizona State
University, Tempe, AZ, USA}
\affil[3]{ASU–SFI Center for Biosocial Complex Systems, Arizona State
University, Tempe, AZ, USA}
\affil[4]{Santa Fe Institute, Santa Fe, NM, USA}

\affil[*]{jake.hanson@asu.edu; sara.i.walker@asu.edu}

\begin{abstract}
The scientific study of consciousness is currently undergoing a critical transition in the form of a rapidly evolving scientific debate regarding whether or not currently proposed theories can be assessed for their scientific validity. At the forefront of this debate is Integrated Information Theory (IIT), widely regarded as the preeminent theory of consciousness because of its quantification of subjective experience in a scalar mathematical measure called $\Phi$ that is in principle measurable. Epistemological issues in the form of the ``unfolding argument'' have provided a concrete refutation of IIT by demonstrating how it permits functionally identical systems to have differences in their predicted consciousness. The implication is that IIT and any other proposed theory based on a physical system's causal structure may already be falsified even in the absence of experimental refutation. However, so far many of these arguments surrounding the epistemological foundations of falsification arguments, such as the unfolding argument, are too abstract to determine the full scope of their implications. Here we make these abstract arguments concrete, by providing a simple example of functionally equivalent machines realizable with table-top electronics that take the form of isomorphic digital circuits with and without feedback. This allows us to explicitly demonstrate the different levels of abstraction at which a theory of consciousness can be assessed. Within this computational hierarchy, we show how IIT is simultaneously falsified at the finite-state automaton (FSA) level (an instance of the unfolding argument) and unfalsifiable at the combinatorial state automaton (CSA) level. We use this example to illustrate a more general set of falsification criteria for theories of consciousness: to avoid being unfalsifiable or already falsified scientific theories of consciousness must be invariant with respect to changes that leave the inference procedure fixed at a particular level in a computational hierarchy. Moving forward, our formalism thereby provides a tight constraint on mathematical theories of consciousness, as well as a more concrete foundation connecting the scientific study of consciousness and computer science.
\end{abstract}
\begin{document}

\flushbottom
\maketitle
\thispagestyle{empty}







\section*{Introduction}
If, and if so how, theories for consciousness can be brought within the purview of science is a subject of intense debate and equally intense importance. The resolution of this debate is necessary for validating theory against experiments in human subjects. It is also critical to recognizing and/or engineering consciousness in non-human systems such as machines. Currently, there is a global, multi-million dollar effort devoted to scientifically validating or refuting the most promising candidate theories, specifically Integrated Information Theory and the Global Neuronal Workspace Theory \cite{reardon2019rival}. At the same time, it is becoming increasingly unclear whether these theories meet the required scientific criteria for validating them. 

Since the early 1990s, scientific studies of consciousness have primarily focused on identifying spatiotemporal patterns in the brain that correlate with what we intuitively consider to be conscious experience. This is due in large part to advances in medical imaging such as electroencephalograms (EEG) and functional magnetic resonance imaging (fMRI) that assess brain activity during different functional behaviors (e.g., sleeping, verbal reports, etc.). The empirical data that results from such tests provide evidence for links between spatiotemporal patterns and inferred conscious states. These links, known as Neural Correlates of Consciousness (NCCs), are well-established and form the basis for an entire subfield of contemporary neuroscience \cite{rees2002neural,metzinger2000neural}. Despite the success of NCCs, however, there is an underlying epistemic issue with the scientific study of consciousness because conscious states are never directly observed within the NCC framework. Instead, they must be \textit{inferred} based on our own phenomenological experience. For example, when a person is asleep we infer they are less conscious than when they are awake because we have first-hand subjective experience of what it is like to be both asleep and awake.

While the epistemic issues associated with NCCs are widely known and discussed, the debate around the possibility of falsifying some of the leading theories of consciousness has recently intensified. This resurgence of interest in what constitutes a valid theory for consciousness is primarily due to the new formalization of the scientific issues in the form of ``unfolding'' arguments  \cite{kleiner2020falsification, hanson2019, doerig2019}. In particular, the original unfolding argument as clarified by Doerig et al. points to deep logical problems with any causal structure theory (CST) that assumes consciousness supervenes on a particular causal structure independent of outward functional consequences \cite{doerig2019}, which implies NCCs would be inadequate to validate such theories. Since the currently leading candidate theory for consciousness, Integrated Information Theory (IIT), is itself a causal structure theory, this has major implications for how we approach the problem of consciousness. To understand how the unfolding argument aims to falsify IIT, it is important to first understand how IIT is constructed as a theory that is derived from simple axioms that make assumptions regarding what conscious experience is, and from these derives a mathematical measure of integrated information $\Phi$ that is proposed as a quantification of consciousness. Among the axioms of the theory is the integration axiom, which states that we experience consciousness as an "undivided whole", meaning, for example, that our left and right visual field are integrated into a single conscious experience. Crucially, integration (and the other phenomenological axioms of IIT) must have a direct translation in terms of mathematical machinery to construct the formal theory. For integration, this is achieved by enforcing integration of the physical substrate(s) that gives rise to consciousness, where the precise mathematical definition is in terms of the presence of feedback between the physical components in a system (e.g., neurons). Consequently, any system that is strictly feed-forward is unconscious, by definition in IIT, due to an assumed inability for such physical structures to generate a unified subjective experience. What the unfolding argument showed was that the input-output behavior of any conscious system with feedback and $\Phi > 0$ can be perfectly emulated by a strictly feed-forward system with $\Phi= 0$. To do so, one simply needs to "unfold" the feedback present in the causal structure of the conscious system in a way that preserves the underlying functionality of the system (i.e. the input-output behavior) - a feat that can be accomplished in the forward or backward direction using feed-forward and recurrent neural networks, respectively \cite{doerig2019} or Krohn-Rhodes decomposition \cite{hanson2019}. The unfolding argument highlights a key issue with IIT and other potential CSTs: the physical process that is assumed to be causally responsible for generating consciousness does not necessarily correlate with any particular input-output behavior, meaning it is not possible to directly test predictions from the theory. The most recent contribution to this debate has come in the form of its generalization by Kleiner and Hoel that applies to any theory for consciousness where the inference from a given measurement of the state of consciousness does not match the prediction under substitution of one physical system for another with appropriate constraints on the substitution \cite{kleiner2020falsification}.

Arguments by Doerig et al. and Kleiner and Hoel have addressed the epistemic issues surrounding falsification of theories of consciousness in the abstract. Here, we seek to ground these abstract arguments in a concrete, easily visualizable system that allows clear demonstration of their consequences. The key contribution of the current work is to demonstrate how the issue of falsification is related to the level in the computational hierarchy at which one assesses the validity of a theory for consciousness. To do so, we introduce a hierarchy of formal descriptions that can be used to describe a given finite-state machine. We show that the discrepancy between whether IIT is falsified or unfalsifiable ultimately depends on the computational scale at which inference of subjective experience is made. In particular, we construct isomorphic causal structures (digital circuits) designed to operate a simple electronic tollbooth with and without feedback. In light of this isomorphism, we evaluate the falsification of IIT at two levels of computation for this circuit: at the finite-state automaton (FSA) level and the combinatorial state automaton (CSA) level and show how the theory is either unfalsifiable at the CSA level or already falsified at the FSA level. Our case study demonstrates how candidate measures of consciousness must be invariant with respect to changes in formal descriptions that exist below the level of the specified inference procedure if they are to avoid \textit{a priori} falsification. An added consequence is that our approach allows a deep connection between the current debate surrounding formalization of falsification arguments with foundations of computer science. We conclude with a brief discussion regarding what a candidate measure of consciousness that satisfies this constraint might look like, as well as the scope of its applicability.

\section*{Results} \label{results}

\subsection*{Defining Falsification for theories of Consciousness}
Falsification is formally defined as a mismatch between a theoretical prediction and an observation and is essential for a theory to be considered scientific \cite{popper2014conjectures}. The scientific study of consciousness is problematic due to the inability to observe conscious states directly. Instead, they must be \textit{inferred} based on some other empirical observation. Thus, falsification for theories of consciousness must be defined as a mismatch between prediction and inference based on observation rather than prediction and direct observation \cite{kleiner2020falsification}. Consequently, it is possible to disagree as to whether or not a theory of consciousness is falsified due to discrepancies between inference procedures being applied to empirical observations (i.e. the empirical data is the same but the inferences are different), or worse, to selectively choose inference procedures depending on the empirical data.

Consensus agreement can only be achieved for falsification arguments if they are explicitly constructed \textit{with respect to a fixed inference procedure}. That is, if a physical system can be transformed into another physical system in a way that preserves the results from the inference procedure while changing the underlying prediction from the theory then a theory of consciousness is falsified with respect to that inference procedure, as this guarantees a mismatch between prediction and inference for at least one of the physical systems under consideration \cite{kleiner2020falsification}. Indeed, this is exactly what is exploited by Doerig et al in their unfolding argument \cite{doerig2019}: the input-output behavior of a system is fixed and the underlying causal structure is transformed in a way that changes the predicted $\Phi$ value from IIT. If one assumes that the inference procedure takes place at the level of input-output behavior, as the authors argue one should, then the preservation of the input-output behavior fixes the inferred conscious experience and falsifies any and all theories of consciousness that are not invariant with respect to this transformation.

\subsection*{The Computational Hierarchy}
Implicitly, it is typically assumed that inference takes place at the level of input-output behavior (e.g. NCCs). However, this is not the only formal level of description at which inferences can be made, nor is it immediately clear that it is inherently the best. At this point, it is at least plausible that lower-level attributes, such as thermodynamic efficiency, may be part of a valid inference procedure. For this reason, we remain agnostic to the precise level at which an inferences are made and instead focus on explicitly characterizing the spectrum of possibilities. To do this, we introduce the following hierarchy that can be used to describe the behavior of a given computational system, allowing us to precisely identify the computational "level" at which a theory is making inferences and predictions. 

\begin{figure}[ht]
    \centering
    \includegraphics[width=0.95\linewidth]{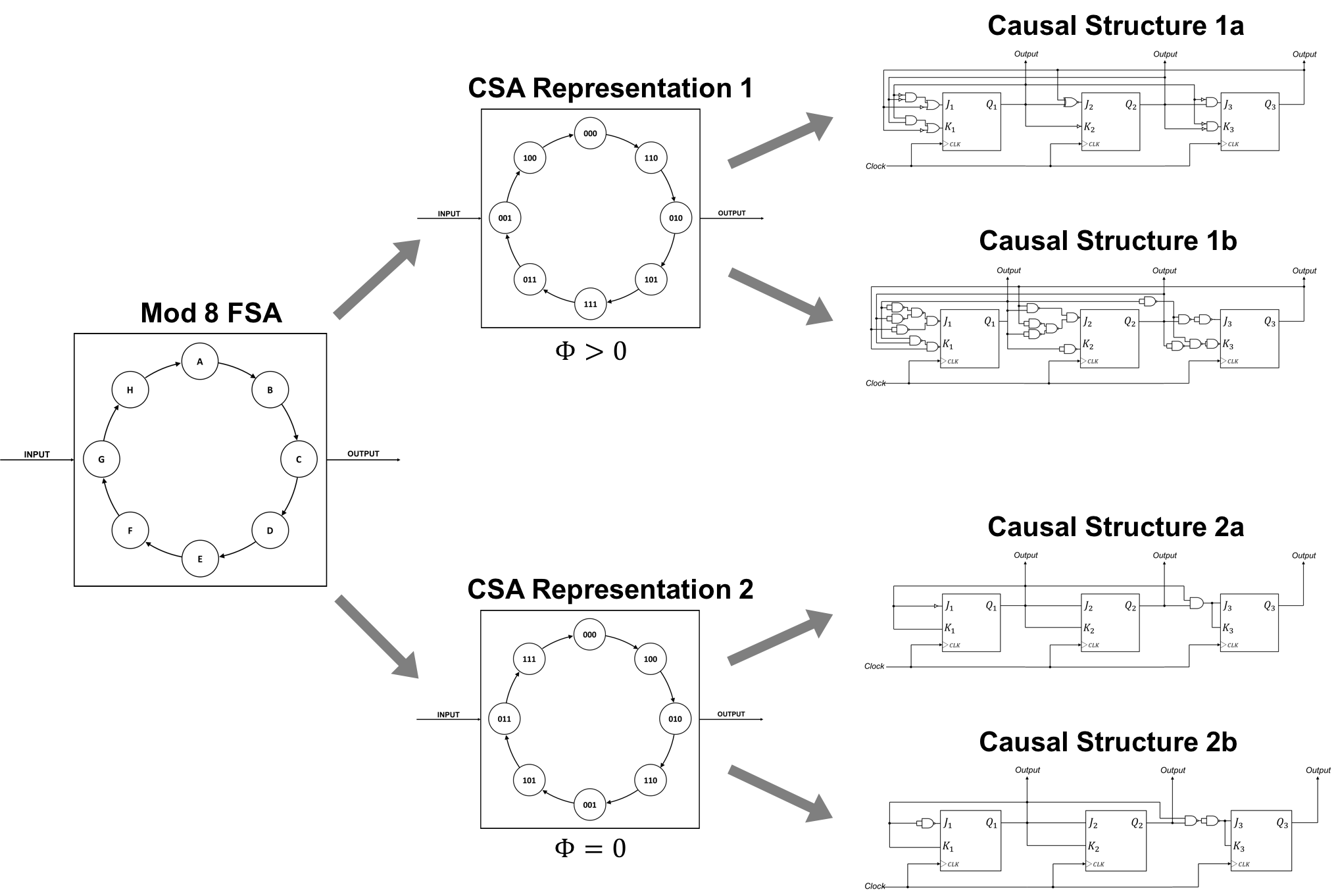}
    \caption{The computational hierarchy used to formally classify levels of inference and prediction. At the top of the hierarchy is the abstract finite-state automaton (FSA) description of a computation which, in this case, is counting mod-eight. Beneath this is the combinatorial state automaton (CSA) description in which abstract states of the FSA have been assigned specific binary labels which, in turn, constrain local dependencies between elements. Note, it is this level of the hierarchy that IIT uses to calculate $\Phi$. At the bottom of the hierarchy is the full causal structure, as specified in terms of the specific logic gates that implement the Boolean functions from the CSA level. In this case, we have shown two different choices for a complete logical basis: \texttt{AND/OR/NOT} gates or universal \texttt{NAND} gates.}
    \label{fig:hierarchy}
\end{figure}

At the top of the hierarchy is the abstract relationship between the inputs, outputs, and internal states that define a computation. These states are typically described in terms of functional behaviors ("stop", "walk", "go", etc.) but what really gives them meaning mathematically is only their topological relationship with one another. This implies that at this level, the formal description of the computation is not grounded in any particular physical representation and could, in fact, be realized by radically different causal structures (Figure \ref{fig:hierarchy}). This abstract treatment of computation corresponds to what Chalmers' refers to as the "finite-state automaton" (FSA) level of description, due to the fact it is defined in terms of a global finite-state automaton \cite{chalmers1993computational}. Beneath this level is what Chalmers refers to as the "combinatorial-state automaton" (CSA) description \cite{chalmers1993computational}. The only difference between the FSA and CSA levels of description is that the latter specifies the computational states of the former in terms of a specific labeling or encoding of the subsystems that comprise the global system. In digital electronics, as well as models of the human brain, this encoding is usually given in terms of binary labels that are assigned to instantiate the functional states of the system. Consequently, transitions between states in the CSA description fix local dependencies between elements, as the correct Boolean update must be applied to each "bit" or "neuron" based on the global state of the system. Furthermore, once a binary representation is specified it constrains the memory required to instantiate the computation, as the number of bits that comprise the system is now fixed. The final level of the hierarchy is the specific choice of logic gates used to implement the Boolean functions specified at the CSA level. This level corresponds to what we would call the "causal structure" as it fully constrains the causal mechanisms that lead to internal state transitions and results in a specific logical architecture (i.e. digital circuit or neuronal wiring). For example, the same Boolean functions (CSA description) can be realized using \texttt{AND},\texttt{OR}, and, \texttt{NOT} gates or universal \texttt{NAND} gates as both form a complete basis for Boolean computation. This choice has interesting physical consequences in terms of the energetic efficiency of a given computation \cite{wolpertcircuits}, though biological systems typically operate many orders of magnitude above the thermodynamic limit \cite{bennett1982thermodynamics}.

\subsubsection*{Prediction and Inference within IIT}
The computational level at which predictions are made in Integrated Information Theory is that of the CSA description, though it is often conflated as a "causal structure" theory. In particular, IIT states that \textit{feedback} between elementary components in a system is a necessary condition for consciousness. The underlying motivation for this assumption is the integration axiom; namely, IIT assumes that an integrated phenomenal experience must be mirrored by integration of the physical substrate that gives rise to consciousness \cite{oizumi2014phenomenology}. In other words, for an experience to be a "unified whole" there must be bidirectional dependencies between the elements that generate this experience. Since the CSA level specifies local dependencies, it is this level that determines the extent to which state transitions rely on feedback between elements and, therefore, the $\Phi$ value for the system. Going below this level is irrelevant, as the dependence between elements is fixed by the Boolean truth tables in the CSA description rather than any particular circuit implementation of these truth tables. Thus, \textit{IIT is invariant with respect to changes below the CSA level}. Consequently, different physical circuits that implement the same CSA (e.g. \texttt{AND/OR/NOT} vs \texttt{NAND} implementations) necessarily have the same $\Phi$ value (Fig. \ref{fig:hierarchy}).

Unlike prediction, there is no clear agreement within IIT as to where in the computational hierarchy one should \textit{infer} conscious experience. In fact, there are clear inconsistencies that are responsible for confusion regarding whether or not IIT is experimentally falsifiable. On one hand, proponents of IIT design experiments to test theoretical predictions against the traditionally held notion that certain outward behaviors such as sleep and self-report are accurate reflections of particular subjective experiences based on our own phenomenal experience. In this case, the inference procedure being used is based on abstract input-output behavior (i.e. the FSA level) where functional states such as sleep are expected in response to inputs such as anesthetics \cite{reardon2019rival, Casali198ra105}. Crucially, none of these states being used for inference have natural binary representations and, therefore, can be encoded in a variety of different ways with a variety of different causal structures. Thus, inferences are made independently of both the CSA and causal structure descriptions in these experiments. On the other hand, proponents of IIT support the claim that it is possible to fix the input-output behavior of a system while still inferring a difference in subjective experiences (i.e. the existence of "philosophical zombies") \cite{oizumi2014phenomenology, Albantakis2019}. In this case, it is the CSA rather than the FSA level of description that must be used to infer the conscious state of a system, as fixed input-output behavior implies a fixed FSA description. Thus, the inference procedure that is used to support the experimental validity of IIT in a traditional laboratory setting must ultimately be rejected in defense of philosophical zombies - a paradox at the heart of the unfolding argument.

\subsection*{A Concrete Example}
\begin{figure}[ht]
    \centering
    
    \begin{subfigure}[t]{0.5\textwidth}
    \includegraphics[width=0.95\linewidth]{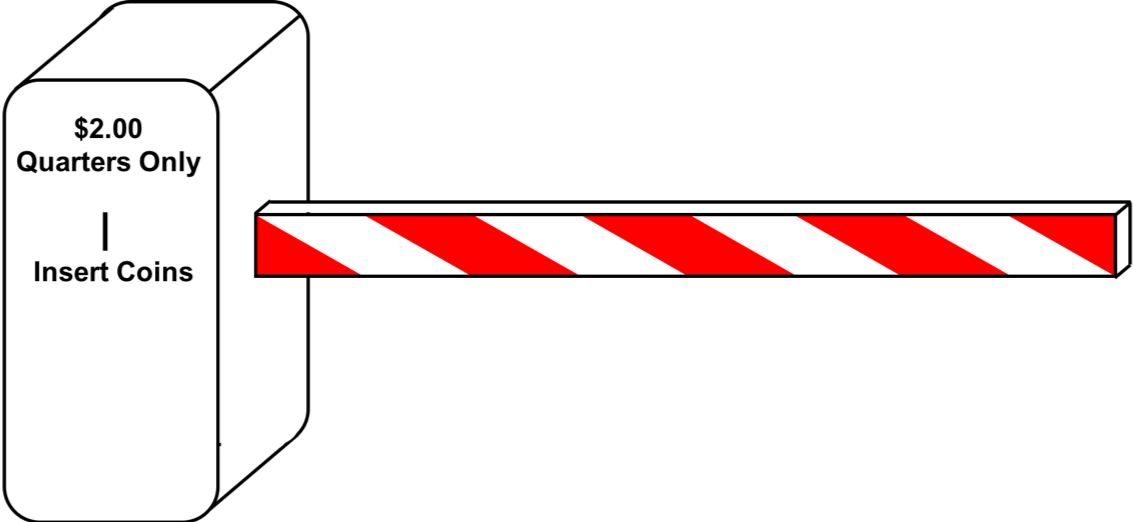}
    \caption{}
    \label{fig:tollbooth_a}
    \end{subfigure}
    \begin{subfigure}[t]{0.4\textwidth}
    \includegraphics[width=0.9\linewidth]{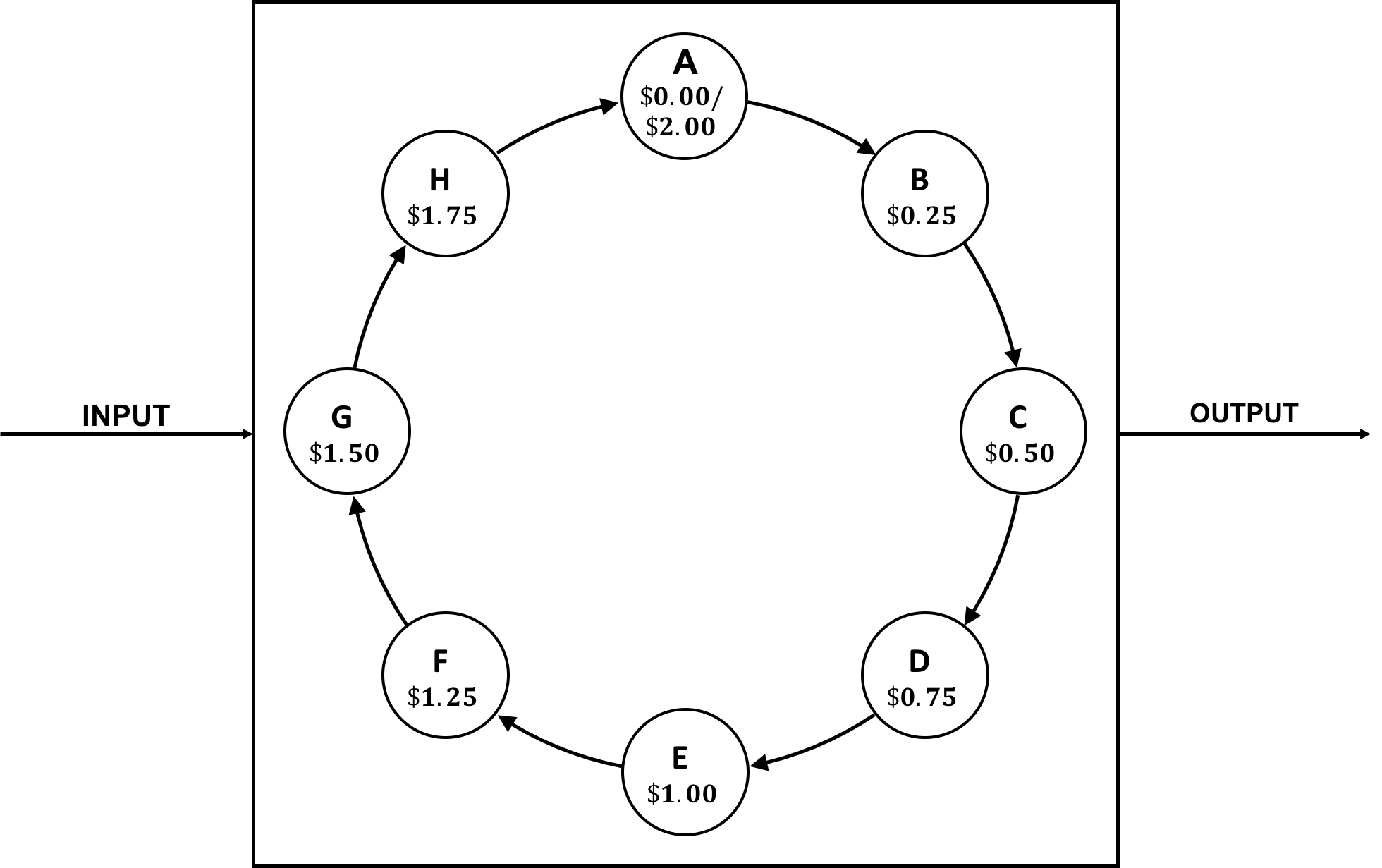}
    \caption{}
    \label{fig:tollbooth_b}
    \end{subfigure}
    
    \caption{Schematic illustration of a simplified electronic tollbooth (\ref{fig:tollbooth_a}) and its FSA description (\ref{fig:tollbooth_b}). The general behavior of the tollbooth is to lift a boom barrier upon receipt of eight quarters ($\$2.00$). To do this requires the ability to cycle through eight internal memory states $\{A, B,..., H\}$, sending each internal state as output to the boom barrier.}
    \label{fig:tollbooth}
\end{figure}

We now turn to a concrete example that demonstrates the logical inconsistencies within IIT, and the more general problem of separating prediction from inference, in terms of easily visualizable tabletop electronics. In particular, we will construct isomorphic digital circuits with and without feedback designed to operate a simple electronic tollbooth, such as that shown in Figure \ref{fig:tollbooth}. Focusing on feedback, as opposed to some other difference in causal structure, allows us to ground our thinking in the specifics of IIT, though the implications of our results readily generalize to any computationalist theory of mind \cite{sep-computational-mind}.

The FSA description of the tollbooth's behavior is defined by the requirement that it must lift the boom barrier in response to the receipt of exactly eight quarters, as shown schematically in Figure \ref{fig:tollbooth_a}. To do this, the circuits governing the behavior of the tollbooth must transition through eight internal memory states, corresponding to the eight functional states in the FSA description of the machine, as shown in Figure \ref{fig:tollbooth_b}. At the CSA level, we insist that both the circuit with feedback and the circuit without feedback be constructed on a three-bit logical architecture, which serves to enforce a strict isomorphism (one-to-one map) between internal states in the two different descriptions. Thus, the FSA description of the system is identical, and the CSA (circuit) descriptions are isomorphic, meaning they instantiate the same \textit{functional} relationship between inputs, outputs, and internal states. Insisting on isomorphic (rather than homomorphic) instantiations allows us to control for all possible confounding factors that can be used to infer a difference in subjective experience, including memory constraints (often referred to as "efficiency" constraints \cite{doerig2019, oizumi2014phenomenology}).

In what follows, we first construct a "conscious" circuit with feedback (and $\Phi > 0$), followed by a functionally identical but "unconscious" circuit with strictly feed-forward connections (and $\Phi =0$). The general construction of both circuits is the same: first, we assign binary labels to the functional states of the system; then, we map these binary state transitions onto JK flip-flops, which are the "bits" in our digital circuits; and last, we use Karnaugh Maps to simplify the logic tables of the JK flip-flops in a way that results in simple elementary logic gate operations (e.g. \texttt{AND}, \texttt{OR}, \texttt{XOR}). As we show, the presence or absence of feedback in the system ultimately stems from the initial choice of the binary labels used to \textit{represent} or \textit{encode} the eight functional states of the system, in accordance with the claim that $\Phi$ acts on the CSA level of description. For the system with feedback, we randomly assign these labels in a way that happens to result in $\Phi>0$ for all states. For the feed-forward system, however, we carefully decompose the underlying dynamics in a way that exploits hierarchical relations such that information flows strictly unidirectionally between components in the system and $\Phi$ is guaranteed to be zero. Note, for the tollbooth to function correctly, the boom barrier must be programmed to recognize the internal state $A$ as functionally important, as this is the output that causes the boom barrier to lift and reset. To avoid confusion over this issue, we simply fix the binary representation of state $A$ as $000$ across CSA representations, corresponding to the notion that the motor hardware of the boom barrier is programmed to recognize this specific binary signal as meaningful. In reality, it is typically assumed that the motor hardware can be reprogrammed to recognize any signal as "meaningful", as all that is relevant from a functional perspective is \textit{consistency} between a circuit and its motor hardware.

\subsubsection*{Constructing a "Conscious" Tollbooth}

To construct the conscious tollbooth, we randomly assign the following binary labels to represent the eight functional states in the FSA description of the tollbooth:
$$
A = 000, B = 110, C = 010, D = 101, E = 111, F = 011, G= 001, H = 100
$$
This assignment of labels fully specifies the CSA description of the system, as each binary component (bit) now must transition in accordance with the current global state of the system. For example, the transition from state $A$ to state $B$ now requires that the first component of the system transitions from binary state $0$ to binary state $1$ when the system is in global state $000$. Similarly, the transition from state $B$ to state $C$ specifies that the first component of the system must transition from $1$ to $0$ when the system is in global state $110$. Taken together, the constraints on each individual component in the system at each moment in time generate a truth table that specifies the interdependencies between elements and, consequently, the $\Phi$ value.

\begin{figure}[ht]
    \centering
    \begin{subfigure}[t]{0.35\textwidth}
    \centering
    \includegraphics[width=0.9\linewidth]{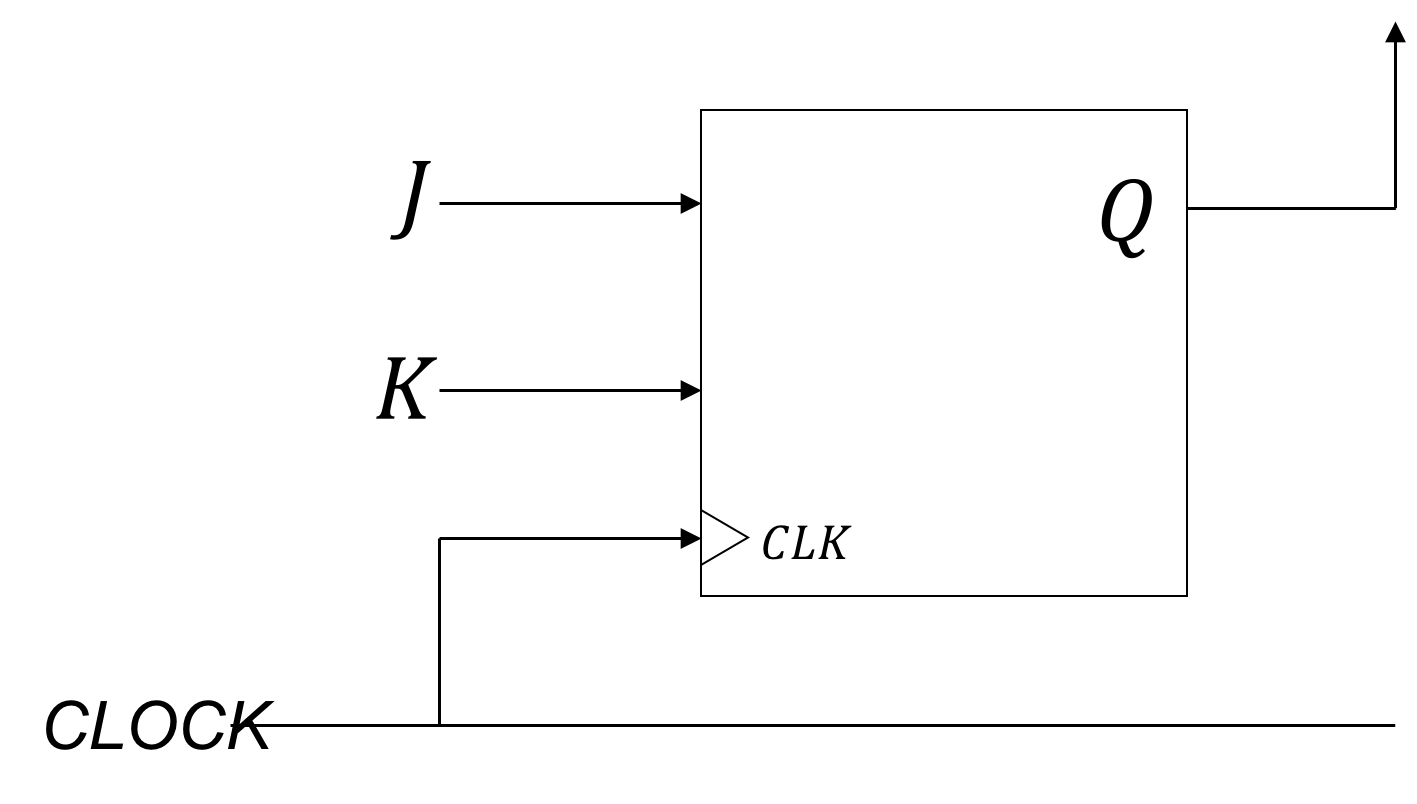}
    \caption{}
    \label{fig:JK_flipflop_img}    
    \end{subfigure}
    \begin{subfigure}[t]{0.55\textwidth}
    \centering
    \includegraphics[width=0.75\linewidth]{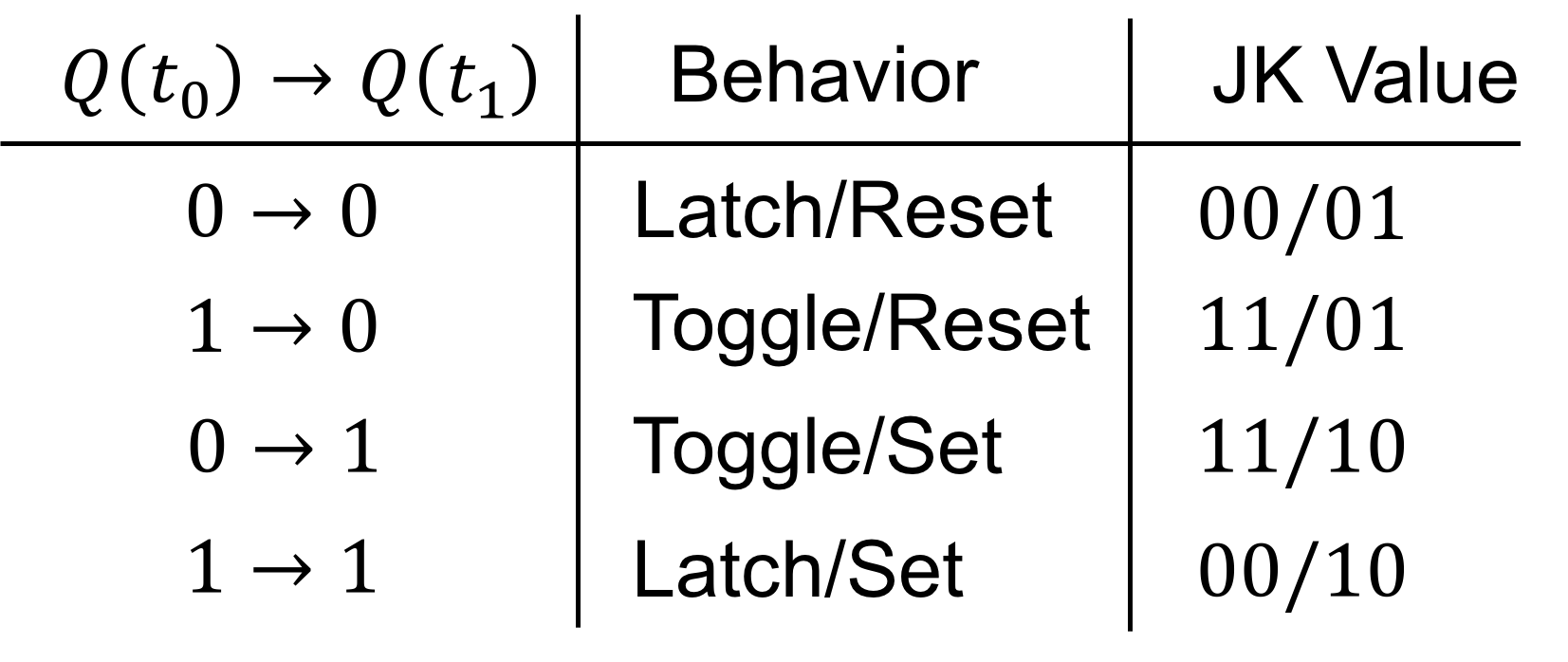}
    \caption{}
    \label{fig:JK_Behavior}       
    \end{subfigure}
    \caption{A JK flip-flop is a widely used binary memory device (bit) in digital electronics (Figure \ref{fig:JK_flipflop_img}). The internal state of the flip-flop takes one of two values ($Q \in \{0,1\}$) and is continuously sent as output. Upon receipt of a voltage from a clocked input, the voltages on the two input channels $J$ and $K$ dictate the state transitions of $Q$ (see main). For any desired internal state transition $Q(t_0) \rightarrow Q(t_1)$, there are two $JK$ inputs that will correctly realize the transition (Figure \ref{fig:JK_Behavior}) which provides flexibility when it comes to circuit design.}
    \label{fig:JK_flipflop}
\end{figure}

To construct the causal architecture, we must specify the elementary building blocks of our system. In a human brain, these building blocks would be neurons but in a digital circuit, these building blocks are "JK flip-flops", which are binary memory storage devices (bits) widely used in the construction of basic digital circuits \cite{moore1958, cavanagh2018sequential}. The behavior of a JK flip-flop is quite simple: there are two stable internal memory states ($0$ and $1$), two input channels (the J input and the K input), and a "clock" that serves to synchronize multiple flip-flops within a circuit. Upon receipt of voltage on a line from the clock, the flip-flop does one of four things depending on the state of the J and K input channel: if the JK input is $00$ the internal state remains unchanged ("latch"), if the JK input is $01$ the internal state resets to $0$ ("reset"), if the JK input is $10$ the internal state is set to $1$ ("set"), and if the JK input is $11$ the internal state is flipped ("toggle"). Thus, for any given internal state transition - $Q_i(t_0) \rightarrow Q_i(t_1)$ - there are two different pairs of JK input that will correctly realize the transition, as shown in Figure \ref{fig:JK_flipflop}. This degeneracy provides flexibility when it comes to the design of the elementary logic gate operations required to actually realize the underlying Boolean logic.

\begin{figure}[ht]
    \centering
    \begin{subfigure}[t]{0.4\textwidth}
    \centering
    \includegraphics[width=0.9\linewidth]{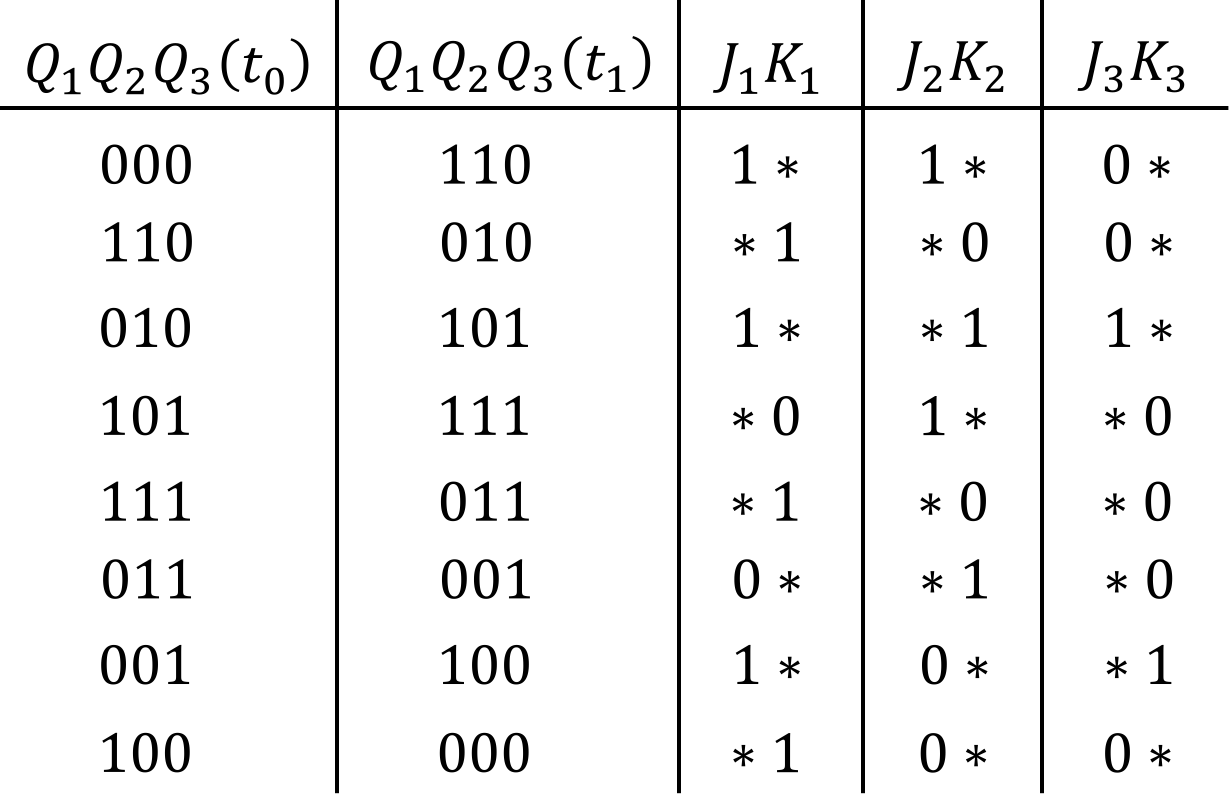}
    \caption{}
    \label{fig:conscious_TPM}    
    \end{subfigure}
    \begin{subfigure}[t]{0.5\textwidth}
    \centering
    \includegraphics[width=0.6\linewidth]{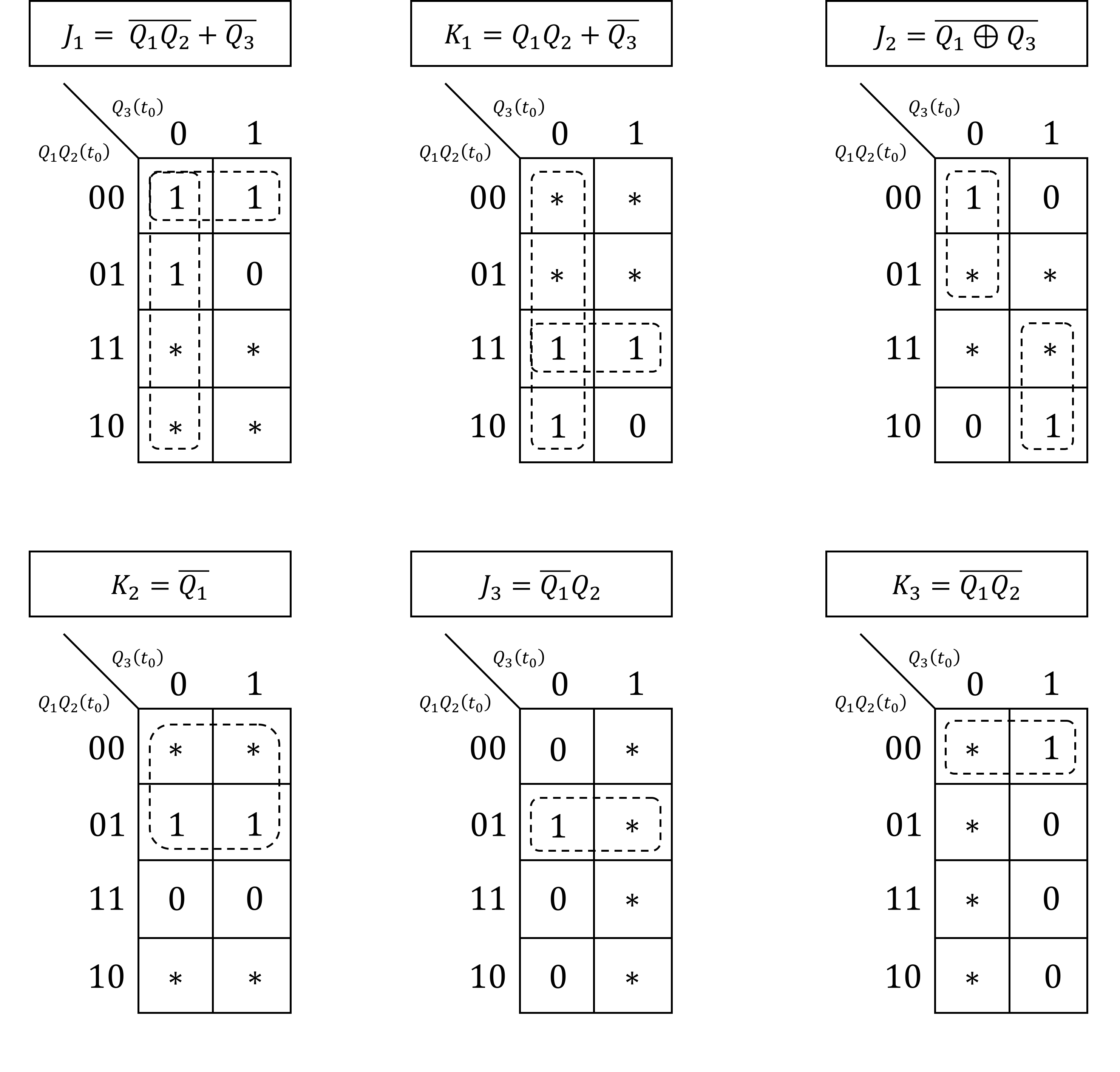}
    \caption{}
    \label{fig:conscious_Kmaps}       
    \end{subfigure}
    \caption{To construct the digital circuit for the conscious tollbooth, we convert the global state transitions into their associated $JK$ values (Figure \ref{fig:conscious_TPM}). Then, we use Karnaugh maps to determine the elementary logic required to update each component (Figure \ref{fig:conscious_Kmaps}). The presence of feedback in the resultant digital circuit is evident by the dependence of earlier components on later components (e.g. $J_1=\overline{Q_1Q_2}+\overline{Q_3}$) and vice versa (e.g. $K_3=\overline{Q_1Q_2}$).}
    \label{fig:conscious_transitions}
\end{figure}

With the specification of the binary labels and the choice of electronic components, we can now finish the construction of the causal structure in terms of elementary logic gates. To do so, we first convert the state transitions of each individual component into their associated JK values. As mentioned, there is degeneracy in the choice of JK input which means we only have to specify one of the input channels (either J or K) to get the desired transition. For each component in the circuit, there is a column in Figure \ref{fig:conscious_TPM} corresponding to the JK value that is required; note, inputs that do not need to be specified are denoted with an asterisk. Next, we must determine the elementary logic gates required to get the correct JK values given the current state of the system. For instance, when the system is in global state $110$, the value of $K_1$ (the K-input to the first component) must be $1$, but when the system is in global state $111$ the value of $K_1$ must be $0$. Taken together, the eight states of the system comprise a truth table of JK input as a function of the global state of the system, as shown in Figure \ref{fig:conscious_Kmaps}. Ordering these truth tables in gray code yields "Karnaugh maps", which allow straightforward identification of the elementary logic gates required to operate the circuit \cite{karnaugh1953}. The elementary logic expression for each of the six input channels, in terms of \texttt{AND},\texttt{OR}, \texttt{XOR}, and \texttt{NOT} gates, is shown above the corresponding Karnaugh map in Figure \ref{fig:conscious_Kmaps}.

\begin{figure}[ht]
    \centering
    \begin{subfigure}[t]{0.84\textwidth}
    \centering
    \includegraphics[width=0.9\linewidth]{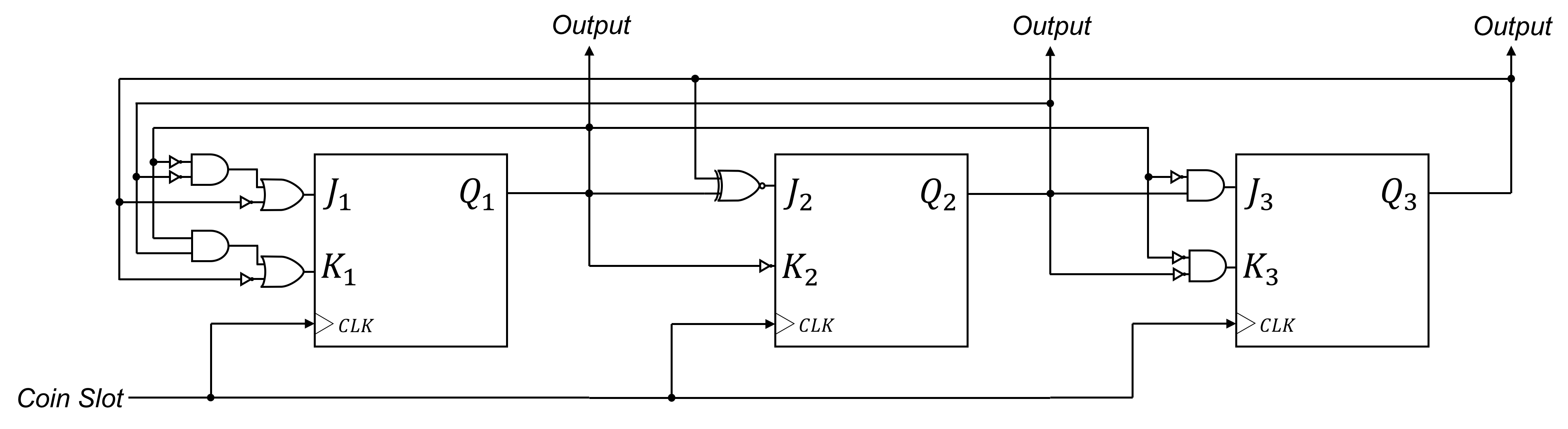}
    \caption{}
    \label{fig:conscious_circuit}    
    \end{subfigure}
    \begin{subfigure}[t]{0.12\textwidth}
    \centering
    \includegraphics[width=1.0\linewidth]{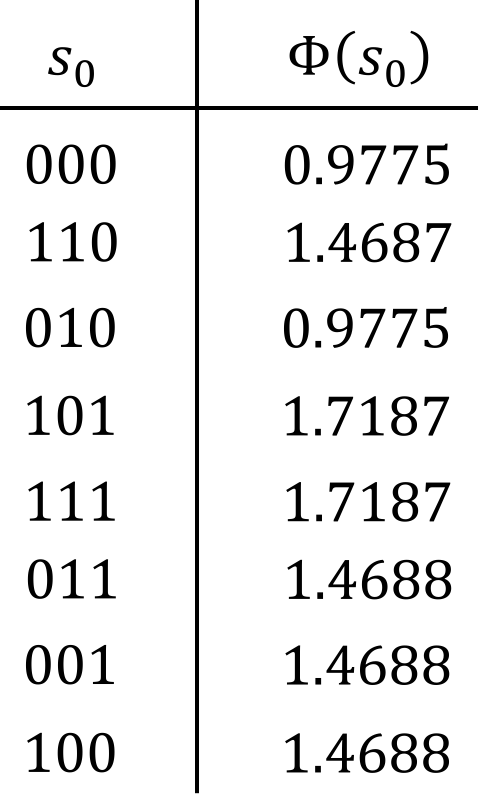}
    \caption{}
    \label{fig:conscious_Phi}       
    \end{subfigure}
    \caption{An integrated digital circuit (Figure \ref{fig:conscious_circuit}) designed to operate the electronic tollbooth shown in Figure \ref{fig:tollbooth}. As can be seen, the causal structure contains meaningful feedback in the form of bidirectional dependencies between pairs of elements and, consequently, has $\Phi>0$ for all states (Figure \ref{fig:conscious_Phi}).
    }
    \label{fig:conscious_entire}
\end{figure}

The elementary logic expressions for the behavior of each JK input completes the construction of our circuit, which is shown in Figure \ref{fig:conscious_circuit}. Clearly, this circuit contains meaningful feedback between components, as the state of the first component depends on the state of the second and third and vice versa. The last thing to check is whether or not this feedback is associated with the presence of consciousness according to IIT, as feedback is a necessary but not sufficient condition for $\Phi>0$. Using the python package PyPhi \cite{mayner2018pyphi}, we find $\Phi>0$ for all states (Figure \ref{fig:conscious_Phi}), meaning this system is indeed considered conscious within the IIT formalism.

\subsubsection*{Constructing an "Unconscious" Tollbooth}
In the previous section, we demonstrated the construction of a causal structure designed to operate the electronic tollbooth shown in Figure \ref{fig:tollbooth_a}. We did so by randomly assigning 3-bit binary labels to represent the function states ($\{A, B,..., H\}$) of the system and constructing the logic of the digital circuit in a way that correctly realizes these labeled state transitions. The result was a circuit that relied on feedback connections (i.e. bi-directional information exchange between components) and had $\Phi>0$ for all states (Figure \ref{fig:conscious_entire}). In this section, we demonstrate that it is possible to assign binary labels in a different way, such that the causal architecture that results instantiates the same functional topology (Figure \ref{fig:tollbooth_b}) but does not make use of any feedback connections. To do so, we will "unfold" the underlying dynamics of the system in a way that guarantees a causal architecture with $\Phi=0$ for all states in the system.

The process of unfolding a finite-state description of a system is based on techniques closely related to the Krohn-Rhodes theorem from automata theory, which states: any abstract deterministic finite-state automata (FSA) can be realized using a strictly feed-forward causal architecture comprised solely of simple elementary components \cite{krohn1965algebraic,zeiger1967yet}. To do so isomorphically, one must find a "nested sequence of preserved partitions", which creates a hierarchical labeling scheme wherein earlier components (flip-flops) transition independently of later components \cite{arbib1968algebraic, hanson2019}. Due to this hierarchical independence, information is guaranteed to flow unidirectionally from earlier components to later components, thereby ensuring a strictly feed-forward logical architecture and, correspondingly, $\Phi=0$ for all states. While a full discussion of Krohn-Rhodes decomposition is beyond the scope of this paper \cite{egri2015computational}, we briefly describe the relevant methodology for constructing a nested sequence of preserved partitions in the Methods section. The result, applied to the finite-state description of the tollbooth shown in Figure \ref{fig:tollbooth_b}, is the following set of binary labels used to encode the functional states of our system:
$$
A = 000, B = 100, C = 010, D = 110, E = 001, F = 101, G= 011, H = 111
$$
Notice, in this labeling scheme, the value of the first component (or "coordinate") partitions the underlying state space of the system into two macrostates: $\{A, C, E, G\}$ and $\{B, D, F, H\}$ and can be thought of as high-level representation of "even" and "odd" states. These macrostates are useful due to the fact they transition deterministically back and forth between one another. Thus, knowing the future state of the first component depends solely on knowing the current state of the first component. Similarly, the future state of the second component is completely deterministic given the current state of the first and second components and is agnostic to the third. In this way, each additional component offers a refined estimate as to where in the global state space the current microstate is located \cite{dedeo2011effective}, hence the claim that the labeling scheme is "hierarchical".

\begin{figure}[ht]
    \centering
    \begin{subfigure}[t]{0.4\textwidth}
    \centering
    \includegraphics[width=0.9\linewidth]{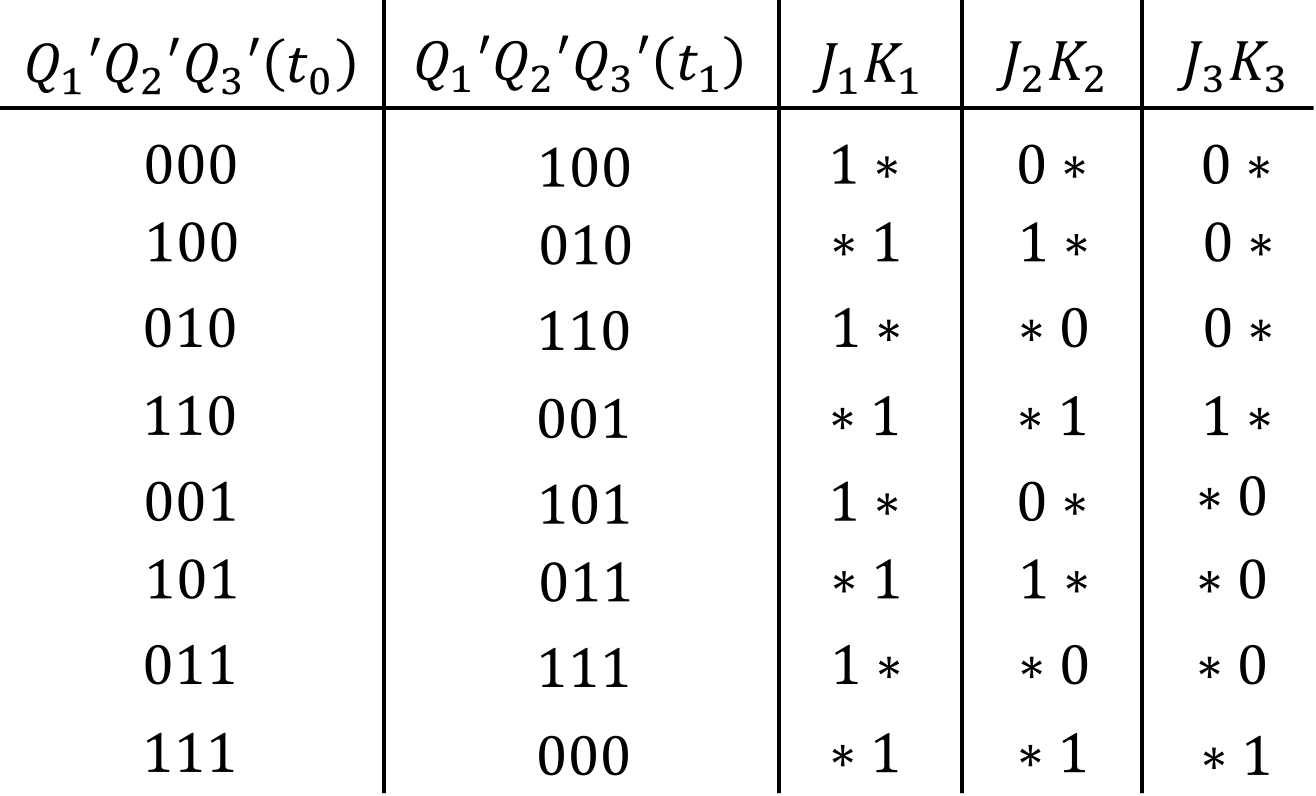}
    \caption{}
    \label{fig:unconscious_TPM}    
    \end{subfigure}
    \begin{subfigure}[t]{0.5\textwidth}
    \centering
    \includegraphics[width=0.6\linewidth]{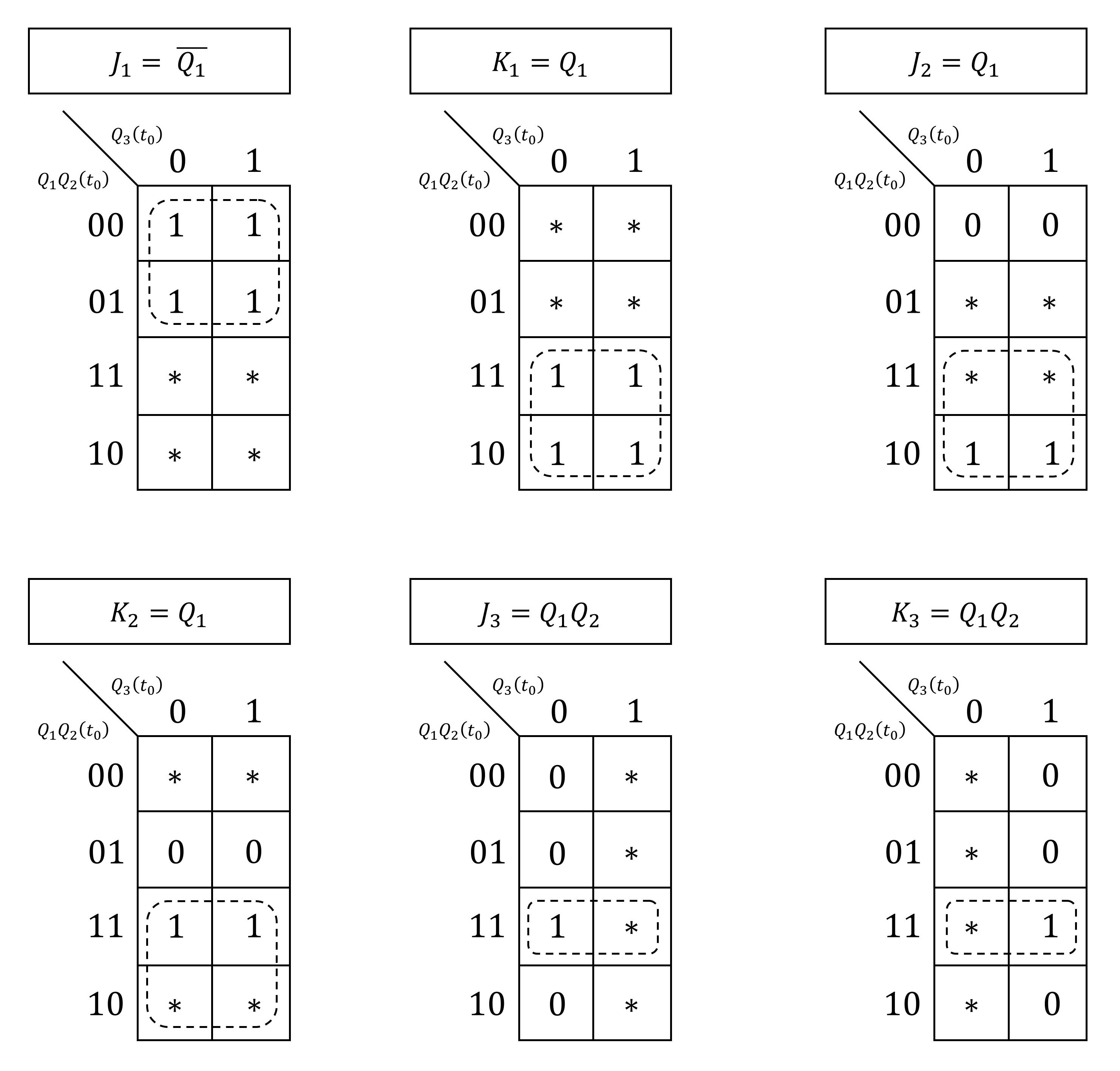}
    \caption{}
    \label{fig:unconscious_Kmaps}       
    \end{subfigure}
    \caption{The state transitions and JK values (Figure \ref{fig:unconscious_TPM}) corresponding to the hierarchical labeling scheme described in the main text. Figure \ref{fig:unconscious_Kmaps} shows the Karnaugh maps used to determine the elementary logic gates used in the construction of the feed-forward logical architecture. Note, the logical dependence between components is strictly unidirectional (e.g. $J_2$ and $K_2$ depend only on the state of $Q_1$).}
    \label{fig:unconscious_transitions}
\end{figure}

\begin{figure}[ht]
    \centering
    \begin{subfigure}[t]{0.84\textwidth}
    \centering
    \includegraphics[width=0.9\linewidth]{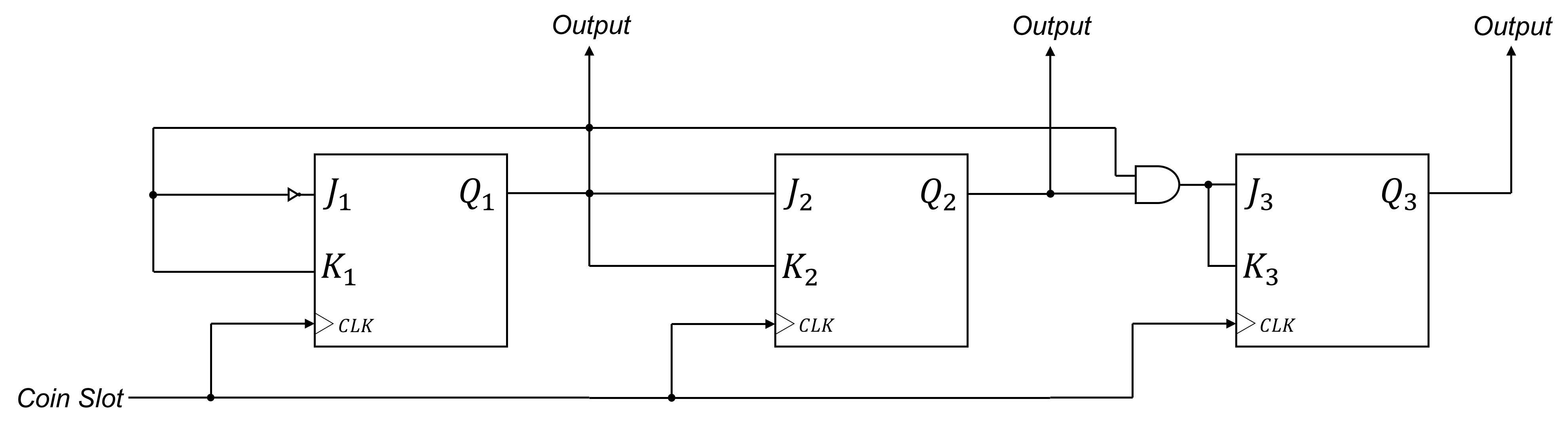}
    \caption{}
    \label{fig:unconscious_circuit}    
    \end{subfigure}
    \begin{subfigure}[t]{0.12\textwidth}
    \centering
    \includegraphics[width=1.0\linewidth]{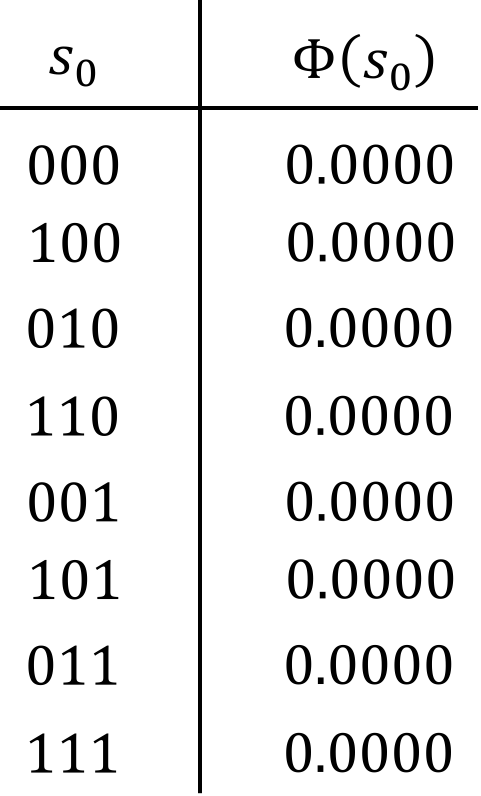}
    \caption{}
    \label{fig:unconscious_Phi}       
    \end{subfigure}
    \caption{A feed-forward digital circuit (Figure \ref{fig:unconscious_circuit}) designed to operate the electronic tollbooth shown in Figure \ref{fig:tollbooth}. This causal structure operates under the same memory constraints as the integrated circuit (i.e. a three-bit logical architecture) but has $\Phi=0$ for all states (Figure \ref{fig:unconscious_Phi}).}
    \label{fig:unconscious_entire}
\end{figure}

With hierarchical labels assigned, the circuit construction now proceeds in a way identical to the previous section. Namely, we convert the binary state transitions into their associated JK values, shown in Figure \ref{fig:unconscious_TPM}. Then, we construct truth tables for the state of each J and K input given the global state of the system; and last, we order these truth tables in gray code (Karnaugh Maps) and assign elementary logic gates to each input channel (Figure \ref{fig:unconscious_Kmaps}). The resulting logical architecture is shown in Figure \ref{fig:unconscious_circuit}. As required, the circuit is strictly feed-forward, as evident by the fact that each component depends solely on itself or earlier components. This, in turn, guarantees $\Phi=0$ for all states of the system (Figure \ref{fig:unconscious_Phi}) as the presence of feedback connections is assumed to be a necessary condition for consciousness according to IIT.

\subsubsection*{Proof of Falsification/Unfalsifiability}
In light of the previous sections, it is clear that IIT \textit{predicts} a difference in subjective experience between the "conscious" tollbooth with $\Phi>0$ and the "unconscious" tollbooth with $\Phi=0$. Thus, falsification is a matter of whether or not one can \textit{infer} a corresponding difference that justifies the difference in prediction. Since the two systems have the same FSA description, any inference procedure that takes place at the FSA level or above necessarily implies falsification of the theory, as the difference in prediction implies a mismatch between prediction and inference for at least one of the two systems under consideration \cite{kleiner2020falsification}. Consequently, IIT is falsified with respect to inference procedures that are based on the input-output behavior of the system, as this is fixed at the FSA level of description.

This implies that the inference procedure utilized by IIT must take place at the CSA level or below if the theory is to avoid total falsification. At the CSA level, however, the full utility of the isomorphism is evident as the only allowable difference between the system with and without $\Phi>0$ was a permutation of the binary labels used to instantiate functional states. Thus, it is this difference that must be used to \textit{infer} a difference in subjective experience. However, unlike input-output behavior, there are no clear phenomenological grounds on which one can infer a difference in subjective experience based solely on a permutation of the binary states used to represent functional states within a system. Instead, IIT must \textit{assume} that such a difference in the CSA description can be used to resolve differences in subjective experience, but it is exactly this assumption that must be tested via comparison of prediction and inference. In other words, if IIT uses the CSA description to infer a difference in subjective experience, then the inference procedure being used is one and the same with the predictions from the theory (i.e. $\Phi$ is used as both inference and prediction), which renders the theory is unfalsifiable. In combination, this implies IIT is falsified with respect to inferences procedures at the FSA level or above and inherently unfalsifiable with respect to inference procedures at the CSA level or below.

\section*{Discussion}

Our results prove an \textit{a priori} falsification of IIT as a scientific theory of consciousness using a simple, readily-realizable model. We have shown that what $\Phi$ actually measures is a consequence of the particular binary representation (or encoding) used to instantiate the functional states in a system at the CSA level, without a clear interpretation in terms of function or phenomenology at the FSA level. For a theory to avoid the epistemic problems revealed by IIT under the isomorphic transformation we introduce requires that no transformation or "substitution" exists that changes the prediction without affecting the inference. This, in turn, implies that beneath the specified level of inference, a mathematical theory of consciousness must be invariant with respect to any and all changes that leave the results from the inference procedure fixed. In other words, if you can make a change to the physical system that does not affect what will be used to infer the conscious state of the system, then such a change must not affect the prediction of the theory either.

For the example we provide, an intuitive measure that satisfies this is Group Complexity \cite{rhodes2009applications}. Like $\Phi$, Group Complexity is a measure of computational complexity that acts on the CSA level of description. Specifically, it counts the number of resets necessary to complete a Krohn-Rhodes decomposition \cite{zeiger1967yet, egri2008hierarchical}, meaning all integrated circuits are decomposed into feed-forward emulations prior to measuring their complexity. This, in turn, puts all CSA representations on an equal playing field, as complexity comes in two forms: "resets" and feedback connections. By first unfolding the dynamics of an integrated circuit, Group Complexity measures the complexity of the underlying computation in the abstract rather than any particular CSA instantiation. Consequently, it is invariant with respect to changes below the FSA level.

In light of this, it is important to ask whether there is anything to be gained from a candidate measure of consciousness such as group complexity. In answer, one must first ensure that the measure is scientific by examining whether inference and prediction can be kept independent. This is easy enough to check for group complexity, as inferences are canonically made based on input-output behavior while group complexity can be measured at the circuit level. Given that there is no a priori dependence between a circuit description and input-output behavior, GC is indeed capable of producing non-trivial scientific predictions. In terms of whether or not these predictions are falsifiable, it is certainly possible that we infer a conscious state based on input-output behavior that is in disagreement with a prediction from a theory based on Group Complexity. For example, if the Group Complexity of a model system increases when the system goes asleep, then this serves as falsification with respect to the canonical inference that sleep should correspond to lower subjective experience. While this may sound virtually identical to experiments designed to test IIT \cite{reardon2019rival, Casali198ra105}, the crucial difference is that group complexity is mathematically invariant with respect to changes that preserve a given FSA description.

Thus, it appears Group Complexity is a measure of complexity that is both non-trivial and falsifiable. Therefore, it is an epistemologically sound measure of consciousness that retains some of the original insight that motivated Integrated Information Theory \cite{tononi1998consciousness} and acts on the same mathematical structures. Yet, at face value, group complexity seems much too simple to truly quantify conscious experience. For one, it coarse grains all of the richness associated with sensorimotor experience into a scalar value that retains none of the corresponding physical information associated with conscious experience, {\it i.e.,} it has no implicit explanation for "what it is like" to be something \cite{Nagel1974}. While IIT deals with this problem by equating multi-dimensional vectors with ``concepts in qualia space", such sophistications are even harder to ground experimentally than a scalar measure, as the ability to empirically resolve the nuances of a rich phenomenal structure are limited by our ability to empirically infer such structures.

Given this, it seems the biggest problem faced by consciousness research going forward is not necessarily the mathematical structures that a theory can \textit{predict} but the mathematical structures that a theory can {\it infer}. We know based on first-hand phenomenal experience of consciousness that certain behaviors such as sleep and verbal report are likely accurate reflections of consciousness in human beings and it is these behaviors that must be leveraged by the inference procedure. Beyond these few specific examples, however, it is difficult to imagine what else can be used to infer conscious states that is not also used to make predictions within the theory. In cases where we lose phenomenological grounding, such as artificial intelligence, this issue is especially problematic \cite{doerig2020empirical}.

While the inability to test what we assume to be consciousness has always plagued the study of consciousness, we hope that formalizing the problem in terms of the level of computational abstraction at which inferences and predictions take place makes it clear that there are mathematical constraints all theories of consciousness must satisfy. Namely, the theory must be invariant with respect to changes that leave the results from the inference procedure unaffected. In IIT, the inference procedure being used to justify the experimental validity the theory is at the level of the input-output behavior of the system, and therefore $\Phi$ must be invariant with respect to equivalence classes that share the same FSA description. The fact that it is not either falsifies the theory or renders it metaphysical, depending on whether or not one accepts the canonical inference procedure. Our analyses indicate that not only are new theories of consciousness needed, but new frameworks for assessing the validity of these theories is needed as well. The latter, for example, could be addressed by constructing theories that do not aim to quantify what subjective experience is, but rather the causal consequences of subjective experience on the physical world.

\section*{Methods} \label{Methods}
\subsection*{Isomorphic Unfolding via Preserved Partitions}
The Krohn-Rhodes theorem guarantees that any finite-state transition diagram can be "unfolded" such that the resultant causal architecure is feedback free and has $\Phi=0$. Typically, however, this unfolding process results in a causal architecture that is much larger than the minimum number of bits to instantiate the functional topology of the system using feedback. In other words, Krohn-Rhodes decomposition, and other unfolding methodologies \cite{oizumi2014phenomenology, doerig2019}, inevitably result in a clear difference in efficiency between feed-forward and recurrent representations of the same underlying computation. To control for this, we must find a system that allows an \textit{isomorphic} feed-forward representation, which can be done using a nested sequence of preserved partitions.

A preserved partition is a way of grouping microscopic states into macroscopic equivalence classes (blocks) based on symmetries present in dynamics. In particular, a partition $P$ is preserved if it breaks the microscopic state space $S$ into a set of blocks $P=\{B_1, B_2, ... , B_N\}$ such that every microstate within a given block transitions to the same macrostate (i.e. the same block) \cite{hartmanis1966algebraic, arbib1968algebraic}. If we denote the underlying microscopic dynamics as a function $f:S \rightarrow S$, then a block $B_i$  is preserved when:
$$
\exists j \in \{1,2,...,N\} \textrm{ such that } f(x) \in B_j \forall x \in B_i
$$
In other words, for $B_i$ to be preserved, $\forall x$ in $B_i$ $x$ must transition to some state in a single block $B_j$ ($i=j$ is allowed). Conversely, $B_i$ is \textit{not} preserved if there exist two or more states in $B_i$ that transition to different blocks (i.e. $\exists$ $x_1,x_2\in B_i$ such that $f(x_1)=B_j$ and $f(x_2)=B_k$ with $j \neq k$ ). In order for the entire partition $P_i$ to be preserved, each block within the partition must be preserved.

For an isomorphic cascade decomposition to exist, we must be able to heirarchically construct preserved partitions in a maximally efficient way. Namely, each partition in the nested sequence of preserved partitions ($\{P_1,P_2,...,P_N\}$) must consist of blocks that evenly split the blocks in the partition above it in half. If this is the case, then a single bit of information can be used to specify where in the preceding block the current state is located. This, in turn, allows a straightforward mapping from the blocks of the preserved partition $P_i$ onto the first $i$ binary coordinates used to represent these blocks. Thus, a system with $2^n$ microstates requires only $n$ binary components, meaning the representation is maximally compact. If one cannot find a preserved partition made of disjoint blocks or the blocks of a given partition do not evenly split the blocks of the partition above it in half, then the system in question does not allow an isomorphic feed-forward decomposition and traditional Krohn-Rhodes decomposition techniques \cite{arbib1968algebraic, egri2008hierarchical, egri2015computational} must be employed.

\begin{figure}[ht]
    \centering
    \includegraphics[width=0.9\linewidth]{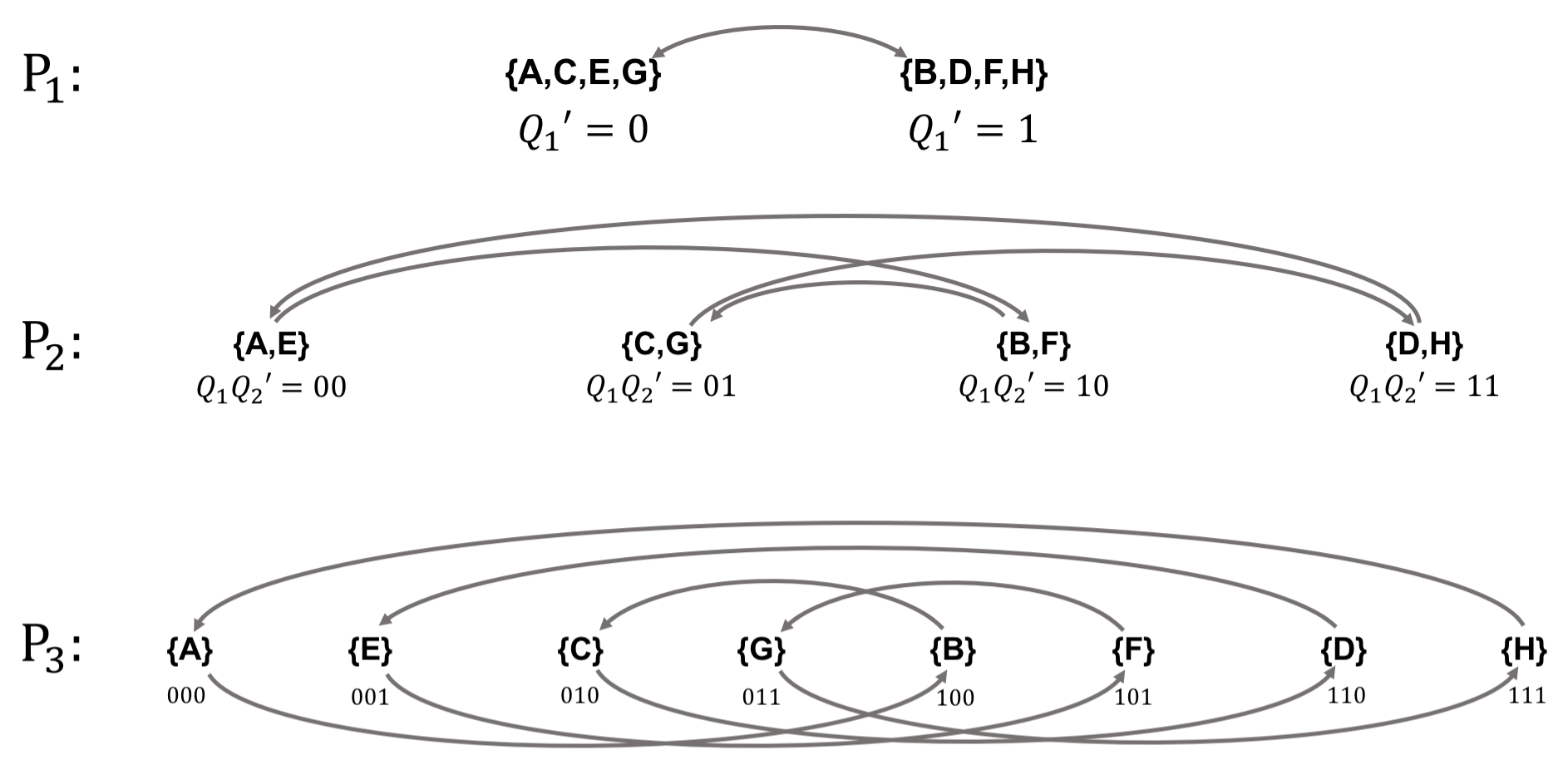}
    \caption{A nested sequence of preserved partitions $\{P_1,P_2,P_3\}$ used to isomorphically decompose or "unfold" the dynamics underlying the finite-state description of the tollbooth shown in Figure \ref{fig:tollbooth}. Blocks within any given partition transition deterministically, which implies the logic for individual components can be constructed hierarchically. The binary labels assigned to the blocks of $P_3$ correspond to a labeling scheme that is isomorphic to the original and strictly feed-forward (see main).}
    \label{fig:preserved_partitions}
\end{figure}

To isomorphically decompose the finite-state automaton shown in Figure \ref{fig:tollbooth_b}, we let our first preserved partition be $P_1=\{B_0,B_1\}$ with $B_0=\{A,C,E,G\}$ and $B_1=\{B,D,F,H\}$. It is easy to check that this partition is preserved, as one can verify that every element in $B_0$ transitions to an element in $B_1$ and every element in $B_1$ transitions to an element in $B_0$ (shown topologically in Figure \ref{fig:preserved_partitions}). To keep track of the blocks, we assign all the states in $B_0$ a binary coordinate value of $Q_1'=0$ and all the states in $B_1$ a binary coordinate value of $Q_1'=1$, which serves as the first of the three binary components ($Q_1'Q_2'Q_3'$) assigned to represent the global state of the system. The logic of the first coordinate is given by the corresponding state transitions of the blocks in $P_1$. Since block $0$ goes to $1$ and vice versa, the first component is essentially a \texttt{NOT} gate taking input from itself, or a JK flip-flop receiving a "toggle" signal.


The second preserved partition $P_2$ must evenly split each block within $P_1$, such that every block in $P_2$ is half the size of the blocks in $P_1$. Denoting $P_2 = \{\{B_{00},B_{01}\},\{B_{10},B_{11}\}\}$, we let $B_{00}=\{A,E\}$, $B_{01}=\{C,G\}$, $B_{10}=\{B,F\}$, and $B_{11}=\{D,H\}$. One can quickly check that these blocks are indeed preserved, and that the component logic for $Q_2'$ (based on the state of $Q_1'Q_2'$) is given by: $\{ 00\rightarrow 0; 01 \rightarrow 1; 10 \rightarrow 1; 11 \rightarrow 0\}$. In a single-channel input scheme, this corresponds to $Q_2'$ as an \texttt{XOR} gate (i.e. $Q_2'=Q_1'\oplus Q_2'$ but, again, the two channel logic corresponding to a JK flip-flop will differ slightly.

The third and final partition $P_3$ must also split the blocks of $P_2$ in half, which implies each of the eight states corresponds to its own block in $P_3$. Naturally, this partition is preserved since there is only a single state in each block (making it impossible for two states within a given block to transition to separate blocks). Since $P_3$ is at the bottom of the hierarchy, the state of $Q_3'$ can depend on the global state of the system ($Q_1'Q_2'Q_3'$). Unlike the previous two coordinates, this truth table is too large to be captured with a single elementary logic gate (e.g. \texttt{NOT},\texttt{XOR},etc.). Instead, we must rely on a combination of elementary logic gates, which is drastically simplified by the use of JK flip-flops. Indeed, it is this third coordinate (and the potential for more complicated logical descriptions in general) that motivated our use of two channel flip-flops rather than single channel devices (e.g. D flip-flops). Reading the block transitions off of the bottom of Figure \ref{fig:preserved_partitions}, we have $\{000 \rightarrow 0; 001 \rightarrow 0; 010 \rightarrow 1; 011 \rightarrow 1;  100 \rightarrow 0; 101 \rightarrow 0; 110 \rightarrow 1; 111 \rightarrow 1\}$. Clearly, there is no single binary logic gate that implements this truth table, and we must instead refer to the Karnaugh maps shown in Figure \ref{fig:conscious_Kmaps}.

At this point, the isomorphic cascade decomposition is complete. The values assigned to the blocks of $Q_3$ correspond to our new binary labeling scheme, namely:

$$
A = 000, B = 100, C = 010, D = 110, E = 001, F = 101, G= 011, H = 111
$$

As demonstrated in the main text, these labels result in a causal architecture that is strictly feed-forward and has $\Phi=0$ for all states, as desired. This can easily be seen by the fact that the transitions of blocks in any given level of the nested sequence of preserved partitions are fully deterministic without the need to specify lower levels (Figure \ref{fig:preserved_partitions}). Thus, downstream information from later coordinates is inconsequential to the action of earlier coordinates, which enforces the "hierarchical" relationship between components. Note, this result is by no means unique; there are other nested sequences of preserved partitions for this system that are equally valid. Choosing a different nested sequence of preserved partitions simply amounts to changing the labels assigned to each block which, in turn, changes the Boolean logic governing the system. As long as the partitions are preserved, however, the causal architecture that results is guaranteed to be strictly feed-forward and isomorphic to the logical architecture we present.

\bibliography{references}






\end{document}